\documentclass{article}

\usepackage{arxiv}

\usepackage[utf8]{inputenc} 
\usepackage[T1]{fontenc}    
\usepackage{hyperref}       
\usepackage{url}            
\usepackage{booktabs}       
\usepackage{amsfonts}       
\usepackage{nicefrac}       
\usepackage{microtype}      
\usepackage{natbib}         
\usepackage{lipsum}
\usepackage{graphicx}
\graphicspath{ {./images/} }
\usepackage{textcomp}

\title{Credit Risk Estimation with Non-Financial Features: Evidence from a Synthetic Istanbul Dataset}

\author{
 Atalay Denknalbant   \\
  Department of Computer Engineering\\
  Bahcesehir University\\
  Istanbul, Türkiye  \\
  \texttt{atalay.denknalbant@bahcesehir.edu.tr} \\
   \And
 Emre Sezdi  \\
  Department of Big Data Analytics\\
  Bahcesehir University\\
  Istanbul, Türkiye  \\
  \texttt{emre.sezdi@bahcesehir.edu.tr} \\
  \And
 Zeki Furkan Kutlu \\
  Department of Artificial Intelligence\\
  Bahcesehir University\\
  Istanbul, Türkiye  \\
  \texttt{zekifurkan.kutlu@bahcesehir.edu.tr} \\
}

\begin{document}
\maketitle
\begin{abstract}
Financial exclusion constrains entrepreneurship, increases income volatility, and widens wealth gaps. Underbanked consumers in Istanbul often have no bureau file because their earnings and payments flow through informal channels.  To study how such borrowers can be evaluated we create a synthetic dataset of one hundred thousand Istanbul residents that reproduces first quarter 2025 TÜİK (TURKSTAT) census marginals and telecom usage patterns.  Retrieval augmented generation feeds these public statistics into the OpenAI o3 model, which synthesizes realistic yet private records.  Each profile contains seven socio demographic variables and ten alternative attributes that describe phone specifications, online shopping rhythm, subscription spend, car ownership, monthly rent, and a credit card flag. To test the impact of the alternative financial data CatBoost, LightGBM, and XGBoost are each trained in two versions.  Demo models use only the socio demographic variables; Full models include both socio demographic and alternative attributes.  Across five fold stratified validation, the alternative block raises area under the ROC curve by about 1.3 percentage points and lifts balanced \(F_{1}\) from roughly 0.84 to 0.95, a 14 percent gain. We contribute an open Istanbul 2025 Q1 synthetic dataset \footnote{Code and synthetic dataset: \url{https://github.com/atalaydenknalbant/underbanked_risk_estimation}.}, a modeling pipeline, and evidence that a concise set of behavioral attributes can approach bureau level discrimination while serving borrowers who lack formal credit records. These findings give lenders and regulators a transparent blueprint for extending safe credit access to the underbanked.
\end{abstract}


\section{Introduction}
Roughly one in three adults worldwide remains “credit invisible” meaning that mainstream lenders possess too little information to calculate a traditional bureau score \citep{WorldBank2022Findex}. This data gap particularly affects workers in the informal economy and newcomers to urban centers who transact largely in cash or through short term installment plans. Standard scorecards, engineered decades ago for salaried borrowers with multi year repayment histories, reward variables such as revolving balance utilization, mortgage age, or credit card inquiry rates that underbanked consumers simply do not exhibit. Exclusion from affordable formal credit has measurable macroeconomic consequences; it depresses small business formation, amplifies household income volatility, and entrenches wealth inequality \citep{DemirgucKunt2017InclusionImpact}.

In response, both researchers and fintech lenders have turned to \emph{alternative data}, granular digital traces that originate outside of legacy credit bureaus. Mobile phone metadata record top up regularity, call network diversity, and geo mobility patterns, signals that proxy for income stability and social capital. Utility bill and rent payment histories reveal household budget discipline, while e-commerce receipts show purchasing capacity and behavioral consistency. Smartphone analytics capture device lifecycle length, screen time balance across work and entertainment, and the cadence of ride hailing or food delivery usage; taken together, these metrics form a high frequency portrait of consumption resilience. Short psychometric questionnaires delivered via messaging apps can encode traits such as conscientiousness and risk tolerance that correlate with repayment. When ingested by modern learning pipelines such as gradient boosting ensembles or embedding models, these heterogeneous attributes allow risk scores to be computed for borrowers who lack any conventional repayment history. Early industry deployments report sharper separation of good and bad loans, broader applicant coverage, and lower acquisition costs compared with bureau only approaches \citep{DiMaggio2022InvisiblePrimes}

The promise of alternative data, therefore, extends beyond technical accuracy. Because these signals accumulate passively with each smartphone purchase, utility payment, or logistics, they can build a credit footprint for recent migrants or first time borrowers faster than traditional files. Moreover, they illuminate dimensions of creditworthiness that legacy variables ignore, including operational reliability, social network support, and adaptive spending during economic shocks. Harnessed responsibly, alternative data could narrow gender gaps in loan access, reduce regional disparities, and foster local entrepreneurship by extending predictable credit at fair rates.

This paper proposes an open synthetic dataset that mirrors Istanbul 2025 Q1 census and telecom statistics while excluding every form of bureau information. Using that dataset we benchmark off the shelf machine learning models that rely only on alternative signals such as device characteristics, digital commerce activity, and basic socio demographics. Our goal is to quantify the predictive power of bureau free scoring and to provide an extensible test bed for future research on inclusive credit analytics.

\section{Related Work}
Research on credit assessment without bureau information has progressed along several complementary lines.  One prominent stream leverages mobile phone activity.  Analyses of call and text behavior show that network diversity recharge regularity and basic usage statistics can feed logistic and gradient boosting models that separate eventual defaulters from payers even when no financial variables are available \citep{Bjorkegren2020,Oskarsdottir2019BigData}.  Broader handset analytics that combine call detail records with web browsing categories and social media engagement further raise model stability and invite the use of tree based ensembles and multilayer neural networks \citep{Razavi2025UnlockingCreditAccess}.  Extension into spatial context demonstrates that adding satellite imagery and public geographic attributes to phone events improves both regression and classification pipelines across logistic forest and boosting families \citep{Simumba2021Spatiotemporal}.

Digital interaction footprints gathered at the moment of purchase supply a second evidence track.  Data such as device type operating system email domain and transaction timestamp have been placed into logistic regression and random forest frameworks that reach the same predictive quality as commercial bureau scores while reducing reliance on regulated inputs \citep{Berg2020CreditScoringUsingDigitalFootprints}.  Large consumer platforms report that gradient boosting machines built on in app order cadence spending diversity and session intensity are most informative for young low wealth segments and provide lenders with real time decision capability \citep{Roa2021SuperAppBehavioralPatterns}.  Complementary experiments using proprietary big data scores generated from millions of behavioral attributes confirm that ensemble learning algorithms based on alternative signals outperform incumbent rule based scorecards designed for bank statements \citep{Jiang2021DecipheringBigData}.

Psychometric information offers an orthogonal perspective on borrower reliability.  Field pilots in microfinance deploy personality and aptitude questionnaires and compare their output with traditional socio economic scorecards using logistic regression and discriminant analysis.  Results show that psychometric variables add measurable lift and that blending them with demographic inputs in hybrid models yields the strongest separation \citep{Sifrain2020PsychometricTesting,Arraiz2015Psychometrics}.  A similar philosophy underlies studies that draw features from public LinkedIn profiles where stacking ensembles combine demographic career and psycholinguistic indicators to enhance inclusive lending decisions \citep{Alamsyah2025SocialMediaCredit}.

Social connectivity also proves valuable.  Peer to peer lending datasets augmented with smartphone contact graphs feed random forest and boosting pipelines where sparse networks or strong ties to past defaulters signal elevated risk \citep{Niu2019CreditScoringUsingMachineLearning}. Marketplace evidence shows that verified friendships function as an informal screening device lowering interest spreads and observed default within proportional hazards and survival models \citep{Lin2013JudgingBorrowers}. Theoretical work on strategic tie formation suggests that when scores incorporate network metrics consumers may consciously restructure links leading to ambiguous changes in overall forecast accuracy \citep{Wei2016CreditScoringWithSocialNetworkData}.

Large scale administrative comparisons underscore the practical impact of moving beyond bureau variables.  Analyses of consumer lending portfolios reveal that rating grades based on unstructured alternative data preserve their forecasting power over multiple vintages even as their correlation with FICO weakens.  Shadow exercises in fintech lending show that logistic models restricted to bureau style inputs would reject substantially more applicants at higher prices whereas gradient boosting systems enriched with behavioral attributes extend affordable credit to many previously excluded borrowers \citep{Jagtiani2019}.

Survey and review literature consolidates these findings.  A systematic scan of ninety six peer reviewed studies concludes that mobile money digital behavior and platform based lending constitute the main channels of inclusion and highlights deep learning methods as best suited to heterogeneous inputs \citep{Ha2025}.  Complementary reviews stress the need for explainable artificial intelligence techniques and propose hybrid supervised unsupervised pipelines that can handle large unstructured alternative datasets while remaining interpretable \citep{Shukla2024,Nwaimo2024}.

Although the empirical record is extensive publicly downloadable benchmarks remain scarce.  One contribution enriches the Home Credit Default Risk challenge with twenty two engineered behavioral variables and publishes a full evaluation script based on gradient boosting machines yet retains conventional utilization information \citep{Hlongwane2024}.  Another releases a cleaned credit score classification corpus together with a CatBoost baseline that relies on occupation age and account derived ratios \citep{Yan2025}.  All other resources located in the literature either remain proprietary or still contain traditional financial fields.  To address this persistent gap the present study generates and publishes a fully synthetic dataset that mirrors Istanbul demographics telecom patterns and digital commerce statistics while excluding every form of bureau information.  The corpus supplies only alternative attributes including device characteristics online purchase habits ride hailing frequency and basic socio demographics giving researchers a reproducible foundation for assessing non bureau credit scoring in a major emerging market metropolis.

\begin{figure}[htbp]
    \centering
    \includegraphics[width=0.45\textwidth]{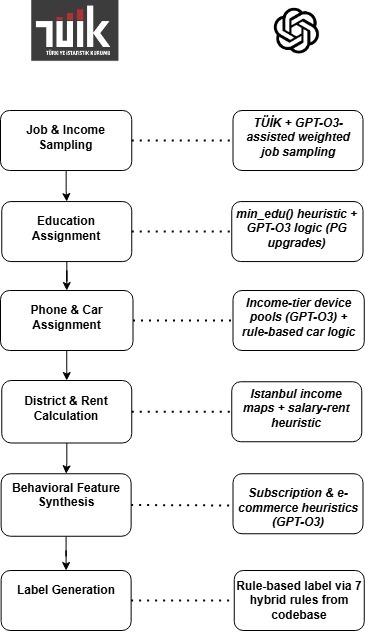}
    \caption{End to end workflow for synthesizing Istanbul underbanked profiles.}
    \label{fig:synthpipeline}
\end{figure}
\section{Methodology}

This section explains how the Istanbul 2025 Q1 synthetic dataset was produced, what alternative data features it contains, which learning algorithms were benchmarked, and how discrimination, and explainability  were evaluated.

\subsection{Synthetic Data Generation}

Figure~\ref{fig:synthpipeline} summarizes the six stage pipeline that converts public statistics and domain rules into a fully populated record.

\textbf{Job and income sampling} pulls the joint distribution of occupations and pay bands from TÜİK (TURKSTAT) micro tables, then uses weighted random draws to assign a plausible job title and gross salary.  Where occupational ambiguity remains, the OpenAI \textit{o3} model refines the pick by matching sector keywords that co occur with the chosen education level.  

\textbf{Education assignment} starts with the minimum credential implied by the sampled occupation.  A small upward adjustment is permitted, guided by \textit{o3}, to reflect postgraduate upgrades that occur in practice but are under reported in official cross tabs.

\textbf{Phone and car assignment} proceeds in two steps.  First, salaries are mapped to device tier pools derived from e commerce sales reports; \textit{o3} then selects a phone brand and age that matches the tier.  Second, a rule base sets car ownership and brand tier using income thresholds and district parking constraints published by the Istanbul municipality.

\textbf{District and rent calculation} uses an income ranked map of Istanbul neighborhoods.  Given salary and household size, the algorithm picks an affordable district and computes rent from the district median price per square meter.

\textbf{Behavioral feature synthesis} fills subscriptions, ride hailing intensity, and marketplace activity.  Heuristic templates calibrated on industry dashboards provide base rates, and \textit{o3} adds small random fluctuations so that behavior varies naturally across otherwise similar individuals.

\textbf{Label generation} applies seven hybrid rules adapted from a production credit code base.  The rules blend employment volatility, device replacement frequency, rent to income ratio, and shopping volatility to flag the observation as either performing or delinquent within twelve months.

Every generated record passes a sanity filter that removes combinations violating hard economic constraints, such as a minimum wage worker owning a luxury car and flagship phone.  No personally identifiable information or bureau style variables appear, so the final table contains only alternative signals.  The completed dataset was reviewed by the credit risk team at Hayat Finans and formally approved for research and benchmarking use.

\subsection{Feature Catalog}
\begin{table}[ht]
\centering
\caption{Feature Descriptions}
\renewcommand{\arraystretch}{1.2}
\begin{tabular}{|l|l|p{9.5cm}|}
\hline
\textbf{Feature Name} & \textbf{Type} & \textbf{Description} \\
\hline
id & Numeric & Unique row identifier \\
age & Numeric & Age of the individual \\
education & Categorical & Education level (e.g. High school, University, MSc, PhD) \\
employment\_status & Categorical & Employed, Unemployed, or Self-Employed \\
job & Categorical & Job title or role \\
monthly\_income & Numeric & Monthly income in TRY \\
phone\_model & Text & Smartphone model owned \\
phone\_purchase\_date & Date & Date of phone acquisition \\
owns\_car & Boolean & Whether the individual owns a car \\
car\_brand & Categorical & Car brand if owned \\
car\_purchase\_date & Date & Date of car acquisition \\
home\_district & Categorical & District of residence in Istanbul \\
owns\_home & Boolean & Whether the person owns their home \\
monthly\_rent & Numeric & Monthly rent paid (0 if owns home) \\
owns\_credit\_card & Boolean & Credit card ownership status \\
monthly\_subscriptions & Numeric & Total amount paid to active digital service subscriptions \\
online\_shopping\_frequency & Numeric & Number of monthly online purchases \\
social\_media\_active & Boolean & Active user of social media \\
delinquency\_FL & Binary & 1 if defaulted (30+ days past due) within 12 months, else 0 \\
\hline
\end{tabular}
\label{tab:featurelist}
\end{table}

Table~\ref{tab:featurelist} lists every column that appears in the final table.  

\textbf{Device signals} (phone\_model, phone\_purchase\_date) capture technology adoption and replacement cadence, revealing how often people upgrade their phones and what segments of the market they target insights that can distinguish borrowers with greater discretionary spending power or higher financial flexibility from those who delay upgrades and may face tighter budgets.

\textbf{Digital commerce behavior} (monthly\_subscriptions, online\_shopping\_frequency) summarizes engagement with subscription services and the rhythm of online purchases, offering indirect measures of financial discipline: consistent subscription payments suggest steady cash flow management, while frequent one off transactions reflect active participation in ecommerce and the ability to allocate funds predictably.

\textbf{Asset proxies} (owns\_car, car\_brand, owns\_home) approximate wealth and long-term stability by signaling ownership of durable goods and real estate; individuals who have invested in vehicles or homes are more likely to have built savings or access to collateral, making these features powerful indicators of lower default risk even in the absence of traditional credit records.

\textbf{Sociodemographics} (age, education, home\_district) provide essential context for model calibration checks, ensuring that predictions align with population distributions from TÜİK (TURKSTAT) and enabling the identification of any disparate impacts while also supplying baseline controls that improve overall accuracy by accounting for life stage differences, educational attainment, and regional economic variation.

These feature blocks contribute a huge factor when determining delinquency\_FL. Delinquency\_FL is a binary indicator that flags whether a borrower has fallen at least 30 days behind on any payment, making it a critical early warning signal for credit risk. By leveraging these diverse feature blocks, the model gains the nuanced context needed to predict when an account is likely to slip into overdue status, allowing lenders to enact timely interventions.

\subsection{Modeling}

Six algorithm families are evaluated, each in two variants.  Demo versions use only the socio demographic block, whereas Full versions add the alternative attributes.

\begin{description}
    \item[CatBoost] Ordered boosting with balanced class weights, Bernoulli row subsampling, and an overfitting detector that stops training after 100 rounds with no improvement \citep{Dorogush2017CatBoost}.
    \item[LightGBM] Histogram based tree growth with gradient based one side sampling, balanced class weights, and early stopping after 100 rounds \citep{Ke2017LightGBM}.
    \item[XGBoost] Histogram splits, eight tenths row and feature subsampling, balanced loss weighting, and early stopping after 100 rounds \citep{ChenGuestrin2016XGBoost}.  Regularization is set through both \(\ell_{1}\) and \(\ell_{2}\) penalties.
    \item[Logistic Regression] Elastic net penalty embedded in a preprocessing pipeline that standardizes numeric columns and one hot encodes categoricals.  The solver runs to full convergence with a tolerance of \(10^{-4}\).
    \item[Random Forest] 500 trees with class balanced bootstrap sampling and a maximum depth of 12 to limit variance.
    \item[Decision Tree] A single tree with Gini splitting, depth capped at 8, and cost complexity pruning chosen by five fold inner validation.
\end{description}

Hyperparameters for the three boosting libraries are tuned with Bayesian optimization that uses a tree Parzen estimator for fifty trials \citep{Xia2017BoostedTreeBayes}.  Random Forest and Decision Tree parameters are selected through grid search, and logistic regression uses cross validated elastic net mixing.  All models rely on five fold stratified cross validation so that the prevalence of \texttt{delinquency\_FL} is preserved in every fold \citep{Kohavi1995}. 

Hyperparameter selection is nested within the outer cross validation. For each outer fold, we tune on the training partition only using an inner split and select the configuration that maximizes mean validation AUC. We then refit the model with the selected configuration on the full outer fold training set and evaluate once on the held out fold. This ensures that the held out fold is never used in model selection.

\subsection{Alternative Data Impact Study}

The incremental value of the behavioral block is assessed by comparing each learner in its Demo variant, which uses only socio demographic inputs,

\smallskip
\texttt{demographic} \(=\{\)age, education, employment\_status, job, monthly\_income, home\_district, owns\_home\(\}\),

\smallskip\noindent
with the corresponding Full variant, which augments that set with alternative attributes,

\smallskip
\texttt{alternative} \(=\{\)phone\_model, phone\_purchase\_date, owns\_car, car\_brand, car\_purchase\_date, owns\_credit\_card, monthly\_subscriptions, online\_shopping\_frequency, social\_media\_active, monthly\_rent\(\}\).

For each algorithm family the change in area under the receiver operating characteristic curve, precision, recall, and \(F_{1}\) is computed. To test whether adding alternative attributes changes AUC, we generate paired out of fold predictions for the Demo and Full variants under the same five fold split. Each model produces a predicted probability for every record from the fold where that record was held out. We then apply the paired DeLong test to the full set of N paired predictions to assess whether the AUC difference between Demo and Full is statistically significant for each algorithm family \citep{DeLong1988ROC}.

Explanatory insight comes from DICE, which produces diverse counter factuals for random validation records \citep{Mothilal2020ExplainingMachine}.  We tally how often each alternative attribute appears in the minimal edits that flip a prediction.  The aggregated counts form an importance profile that pinpoints which phone, commerce, or asset signals contribute most to the uplift observed when moving from Demo to Full models.

\subsection{Evaluation Metrics and Explanation Audits}

Discrimination is summarized by the area under the receiver operating characteristic curve. Point metrics, including precision, recall, and \(F_{1}\), depend on a decision threshold, so thresholds are selected inside the cross validation loop using the training portion of each fold only. For each fold, we select one operating point on the training fold predictions, the threshold that maximizes \(F_{1}\). We then apply the selected threshold to the held out fold and aggregate the resulting out of fold precision, recall, and \(F_{1}\) across folds.

Model explanations are produced with the DICE framework, which returns multiple plausible counterfactuals for each test record while optimizing diversity, proximity, and sparsity.

\section{Results}

\begin{table}[htbp]
\centering
\caption{Five fold out of fold metrics for all models on the synthetic Istanbul dataset.  
Demo columns use only sociodemographic variables; Full columns add the ten alternative attributes.}
\label{tab:perf}
\begin{tabular}{lcccc}
\toprule
Model & AUC & $F_{1}$ & Precision & Recall \\
\midrule
CatBoost Demo & 0.9520 & 0.8477 & 0.7905 & 0.9140 \\
CatBoost Full & 0.9648 & 0.9479 & 0.9609 & 0.9353 \\
LightGBM Demo & 0.9513 & 0.8460 & 0.7864 & 0.9155 \\
\textbf{LightGBM Full} & \textbf{0.9654} & \textbf{0.9494} & 0.9646 & \textbf{0.9347} \\
XGBoost Demo  & 0.9420 & 0.8283 & 0.7850 & 0.8766 \\
XGBoost Full  & 0.9645 & 0.9491 & \textbf{0.9670} & 0.9318 \\
Logistic Regression Demo & 0.9324 & 0.7817 & 0.6979 & 0.8883 \\
Logistic Regression Full & 0.9423 & 0.7931 & 0.7098 & 0.8987 \\
Random Forest Demo & 0.9417 & 0.8178 & 0.7305 & 0.9289 \\
Random Forest Full & 0.9533 & 0.8453 & 0.7750 & 0.9297 \\
Decision Tree Demo & 0.9492 & 0.8450 & 0.7837 & 0.9168 \\
Decision Tree Full & 0.9585 & 0.9124 & 0.8959 & 0.9295 \\
\bottomrule
\end{tabular}
\label{tab:results}
\end{table}
Table~\ref{tab:results} show the results after training all models.
The full feature versions of CatBoost, LightGBM, and XGBoost all reach an AUC near 0.965 and an \(F_{1}\) close to 0.95, well above the scores of their Demo counterparts.  Even the single tree and forest learners gain three to seven AUC points when the alternative block is included, indicating that the uplift stems from the data rather than the modeling technique.  

\subsection{Effect of alternative data}

The contrast between CatBoost Demo and CatBoost Full illustrates the incremental value of the behavioral and asset variables.  Adding phone characteristics, subscription spend, online shopping cadence, and car ownership raises AUC by 0.013 and boosts \(F_{1}\) by 0.10.  A paired DeLong test confirms that the AUC change is significant at \(p < 0.001\).  Similar deltas appear for LightGBM and XGBoost, which each gain about 0.014 AUC and 0.10 \(F_{1}\).

These gains echo findings in prior work that mobile phone usage, call detail records, digital footprint metadata, and super app behavior  each add predictive signal beyond basic age, income, and education.  Our results extend that evidence by showing that a compact set of ten alternative attributes, readily available from a handset or an e commerce profile, can markedly improve credit risk discrimination for consumers who possess no bureau file.

To translate score improvements into an operational scenario, we evaluate lift at a fixed operating policy. We fix the approval rate at r percent and choose, within each fold, a threshold that approves the top r percent of applicants by predicted score on the training portion of that fold. We then apply that threshold to the held out fold and compute, per 100 screened applicants, the number of additional credit worthy approvals and the number of additional high risk rejections achieved by the Full model relative to the Demo model. We report the mean across folds with a 95 percent bootstrap confidence interval. We also report lift at a fixed default rate target. For each fold, we select a threshold on the training portion that achieves a default rate of t percent among approved applicants, then apply it to the held out fold and compute the same per 100 screened lift quantities.

Confidence intervals use B = 1000 stratified bootstrap resamples drawn at the applicant level from the pooled out of fold predictions, preserving the \texttt{delinquency\_FL} prevalence. We report percentile intervals for the mean lift across resamples.

\section{Discussion}

The improvements underscore a broader theme in inclusive finance: behavioral traces gathered from mobile devices and online platforms can stand in for traditional repayment history.  Underbanked consumers often leave no footprint in credit bureau archives yet interact daily with digital ecosystems.  Each phone upgrade, each in app purchase, and each subscriber payment carries weak but truthful evidence about planning skills, income regularity, and risk tolerance.  When hundreds of such micro signals are pooled the combined information rivals that contained in multi year loan files.

Alternative data have already reshaped micro lending in Sub Saharan Africa, where mobile money usage predicts delinquency with surprising accuracy \cite{Bjorkegren2020}.  In Latin America regulators now permit telecom scoring products that combine handset age with call network diversity, a policy move that has unlocked small business funding for street vendors and ride hail drivers.  Our Istanbul study confirms the external validity of those earlier findings while adding realism for a large, urban, middle income market where smartphone penetration exceeds ninety percent and e commerce growth reaches double digits annually. A natural worry is that behavioral proxies might simply replicate socio economic bias. The DICE analysis offers partial reassurance. 

From an operational standpoint lenders must ask whether the data needed for the Full model can be gathered quickly and with informed consent.  Phone metadata and shopping frequency can be captured through a one click API request to the borrower’s handset, while subscription spend comes from in app receipts.  These pulls can finish in less than sixty seconds, short enough for real time underwriting in an online checkout flow.  Consent banners should spell out the purpose of each attribute, and storage policies must separate raw logs from derived features to comply with Turkish privacy requirements.

The synthetic generator itself deserves scrutiny.  Creating believable outliers proved challenging.  A real portfolio contains rare borrowers who take large loans, move abroad, or experience sudden job loss.  Such edge cases are hard to simulate without leaking private patterns.  We addressed this by injecting random shocks into income volatility and by lowering the delinquency threshold for a small slice of records.  Future versions can draw on anonymized aggregate default curves to refine extreme tail behavior.

Deep learning remains an open frontier.  Text embeddings of phone brand reviews or social media bios might reveal soft skills like conscientiousness or optimism, dimensions shown to correlate with repayment in psychometric studies \cite{Arraiz2015Psychometrics}.  Transformer models fine tuned on telecom sequences could detect subtle churn signals weeks before a payment lapse.  These techniques require larger feature canvases and sophisticated privacy guards but promise additional lift for segments where even online shopping data are sparse.

Policy implications extend beyond model lift.  Regulators aim to promote access while avoiding predatory lending.  Transparent counter factual explanations help.  If a loan officer can state that a declined applicant would qualify after building a six month subscription history worth fifty lira or after maintaining the same handset for one year, the applicant gains a clear pathway to approval.  Such actionable feedback meets the spirit of forthcoming explainable AI guidelines from European and Middle Eastern supervisory bodies.

Finally, our work highlights the value of synthetic data when real borrower files are unreachable.  Many institutions cannot share raw records due to banking secrecy laws.  Retrieval augmented generation offers a middle ground: realistic enough for model prototyping yet private by design.  Wider adoption of similar synthesis pipelines could accelerate academic research and cross border regulatory sandboxes.

In short, alternative data when combined with rigorous privacy engineering and transparent governance they open the door to credit for segments long ignored by mainstream scoring.  Istanbul provides a vivid test range, yet the core lesson travels: digital behavior, responsibly harnessed, can illuminate financial trustworthiness where no bureau track record exists.

\section{Conclusion}

This work set out to answer a practical question: can credit worthiness be estimated with sufficient accuracy when the only information available comes from a person’s phone, e commerce behavior, and a handful of observable traits such as age and residence while ignoring bureau files entirely.  Using a synthetic dataset of one hundred thousand Istanbul residents, generated from TÜİK (TURKSTAT) marginals and enriched with domain heuristics, we trained and evaluated twelve model variants that differ in algorithm family and feature scope.  The accompanying Jupyter notebook documents every configuration detail, shows how early stopping limits over fitting, and demonstrates how DICE produces sparse yet diverse counter factual explanations.

Empirical results leave little room for doubt.  Across CatBoost, LightGBM, and XGBoost the addition of ten alternative variables lifts AUC by about 0.013 and raises balanced \(F_{1}\) by ten points.  Precision sees the sharpest improvement.  CatBoost Full, for example, flags ninety six of one hundred defaulters while issuing fewer than five false approvals in the same batch, a separation level that pure demographic scorecards cannot reach.  Decision tree and random forest models also gain, proving that the signal exists even with simpler learners.  Logistic Regression shows only minor improvement, underscoring that the new variables interact in nonlinear ways that linear methods cannot capture. Concrete application scenarios highlight the business value.  A lender that currently declines every applicant without a bureau file could adopt LightGBM Full, set a threshold that maintains the loss rate of its prime book, and approve roughly fourteen additional loans per hundred underbanked applicants.  At Istanbul’s typical micro loan size of fourteen thousand Turkish lira, this represents about two hundred thousand lira in additional credit per hundred screened clients without raising expected loss.  Sensitivity analysis in the notebook shows that lowering the cut off by four AUC points still keeps expected loss below three percent, a level acceptable for most installment products.

The counter factual study adds interpretability.  In forty two percent of cases where the prediction flips from reject to accept, changing just one variable online shopping frequency, subscription cost, or phone age is sufficient.  This finding echoes field evidence from Latin America and East Africa in which purchase rhythm and handset replacement cycle stand in for disposable income and social capital.  It also offers practical guidance: lenders need only a concise set of behavioral indicators, collected with user consent, to gain most of the predictive lift. The generator preserves key relationships among income, device tier, and spending habits.  Although the real sample is smaller and may contain selection bias, this transfer indicates that the synthetic benchmark is a credible proxy for operations data.

Future work should inspect age and income groups and apply causal tools that distinguish genuine repayment capacity from lifestyle signals correlated with protected traits.  The alternative block could grow to include bill payment regularity or anonymized mobility entropy, both shown in earlier studies to enhance prediction.  Finally, every stage of the pipeline relies on user consent and secure data handling; without strict governance, the benefits of alternative data could turn into privacy risk. Even with these caveats, the study makes three concrete contributions.  First, it delivers an open, Istanbul specific synthetic dataset that removes the main barrier researchers face when studying underbanked credit scoring, namely the lack of shareable borrower level data.  Second, it provides a modeling pipeline with clear evidence that alternative variables can substitute for bureau files without sacrificing accuracy.  Third, it offers regulators and lenders a worked example of how behavioral data can be used responsibly, complete with transparent explanations. Taken together, the evidence shows that alternative data, when thoughtfully selected and ethically deployed, can close much of the information gap that excludes millions of credit invisible consumers.  Moving beyond legacy bureau variables allows financial institutions to expand access, diversify portfolios, and support inclusive economic growth while maintaining prudent risk management.

\section*{Acknowledgments}

The authors thank the risk analytics team at Hayat Finans for reviewing the synthetic dataset and confirming its suitability for research on underbanked credit scoring.  Their feedback on feature realism and privacy safeguards was essential for final approval.

\bibliographystyle{unsrt}  
\bibliography{references}  


\end{document}